\documentclass[a4paper]{article}

\usepackage{xcolor}

\newcommand{\asr}{\mathrm{asr}}
\newcommand{\att}{\mathrm{a}}
\newcommand{\ctc}{\mathrm{c}}
\newcommand{\enc}{\mathrm{e}}
\newcommand{\spk}{\mathrm{spk}}

\usepackage{tikz}
\usepgflibrary{arrows.meta}
\usetikzlibrary{positioning}
\usepackage[tight]{subfigure}
\usepackage{cite}

\usepackage{INTERSPEECH2019}
\usepackage{multirow}
\usepackage[hidelinks]{hyperref}

\title{Privacy-Preserving Adversarial Representation Learning in ASR:\\ Reality or Illusion?}
\name{Brij Mohan Lal Srivastava$^1$, Aur\'elien Bellet$^1$, Marc Tommasi$^{2}$, Emmanuel Vincent$^3$}
\address{
  $^1$INRIA, France $^2$Universit\'e de Lille, France \\
  $^3$Universit\'e de Lorraine, CNRS, Inria, Loria, F-54000 Nancy, France}
\email{\{brij.srivastava, aurelien.bellet, marc.tommasi, emmanuel.vincent\}@inria.fr}

\begin{document}

\maketitle
\begin{abstract}


Automatic speech recognition (ASR) is a key technology in many services and applications. This typically requires user devices to send their speech data to the cloud for ASR decoding. As the speech signal carries a lot of information about the speaker, this raises serious privacy concerns. As a solution, an encoder may reside on each user device which performs local computations to anonymize the representation. In this paper, we focus on the protection of speaker identity and study the extent to which users can be recognized based on the encoded representation of their speech as obtained by a deep encoder-decoder architecture trained for ASR. Through speaker identification and verification experiments on the Librispeech corpus with open and closed sets of speakers, we show that the representations obtained from a standard architecture still carry a lot of information about speaker identity. We then propose to use adversarial training to learn representations that perform well in ASR while hiding speaker identity. Our results demonstrate that adversarial training dramatically reduces the closed-set classification accuracy, but this does not translate into increased open-set verification error hence into increased protection of the speaker identity in practice. We suggest several possible reasons behind this negative result.


  
\end{abstract}
\noindent\textbf{Index Terms}: speech recognition, end-to-end system, privacy,
adversarial training, speaker recognition

\section{Introduction}
\label{sec:intro}

With the emergence of pervasive voice
assistants~\cite{lopez2017alexa,kepuska2018next} like Amazon Alexa,
Apple's Siri and Google Home, voice has become one of the most widespread forms of
human-machine interaction. In this context, the speech signal is
sent from the user device to a
cloud-based service, where automatic speech recognition (ASR) and natural language
understanding are performed in order to address the user request.\footnote{See
e.g., {\scriptsize
\url{https://cloud.google.com/speech-to-text/}}}
While recent studies have identified
security vulnerabilities in these devices 
\cite{lei2017insecurity,chung2017alexa}, such studies tend to hide more
important privacy risks that can have long-term
impact. Indeed, state-of-the-art speech processing algorithms can infer not only the
spoken contents from the speech signal, but also the
speaker's identity~\cite{reynolds1995speaker},
intention~\cite{gu2017speech,hellbernd2016prosody,ballmer2013speech,stolcke1998dialog},
gender~\cite{zeng2006robust,kotti2008gender}, emotional
state~\cite{el2011survey,ververidis2004automatic,kwon2003emotion},
pathological
condition~\cite{dibazar2002feature,umapathy2005feature,schuller2013interspeech},
personality~\cite{schuller2013computational,schuller2015survey} and
cultural~\cite{sekiyama1997cultural,vinciarelli2009social} attributes to a great extent. 
These algorithms require just a few tens of hours of training data to
achieve reasonable accuracy, which is easier than ever to collect via virtual assistants. The dissemination of voice signals in large data
centers thereby poses severe privacy threats to the users in the long run.

These privacy issues have little been
investigated so far. The most prominent studies
 use homomorphic encryption and bit string comparison \cite{pathak2012privacy,glackin2017privacy}.
While these methods
provide strong cryptographic guarantees, they come at a large computational
overhead and can hardly be applied to state-of-the-art end-to-end deep neural network based systems.

An alternative software architecture
is to pre-process voice data on the device to remove some personal information
before sending it to web services.  Although this does not rule out all
possible risks, a change of representation of the voice signal can contribute
to limiting unsolicited uses of data. In this paper, we investigate how much of a
user's \emph{identity} is encoded in speech representations built for
ASR.  To this end, we conduct closed- and
open-set speaker recognition experiments. 
The {\it closed-set} experiment refers to a classification
setting where all test speakers are known at training time. In contrast,
 the {\it open-set} experiment (a.k.a.\ speaker
verification) aims to 
measure the capability of an attacker to discriminate between speakers in
a more realistic setting where the test speakers are not known beforehand.
We implement the attacker with the state-of-the-art x-vector speaker
recognition technique \cite{snyder2018x}.

The representations of speech we consider in our work are given by the encoder
output of end-to-end deep encoder-decoder architectures trained for ASR. Such
architectures are natural in our privacy-aware context, as they correspond
to encoding speech on the user device and decoding in the cloud. Our baseline network follows the ESPnet architecture 
\cite{watanabe2018espnet}, with one encoder
and two decoders: one based on connectionist temporal
classification (CTC) and the other on an attention mechanism.
Inspired by \cite{feutry2018learning}, we further propose to extend the
network with a {\it speaker-adversarial} branch so as to learn
representations that perform well in ASR while hiding the speaker identity.

Several papers have recently proposed to use
adversarial training for the goal of improving ASR performance by making the
learned representations invariant to various conditions. While general
form of acoustic variabilities have been studied \cite{serdyuk2016invariant},
there is some work specifically on speaker invariance 
\cite{tsuchiya2018speaker,meng2018speaker}.
Interestingly, there is no general consensus on whether it is more appropriate
to use speaker classification in an adversarial or a multi-task
manner, despite the fact that these two strategies implement opposite means 
(i.e., encouraging representations to be speaker-invariant or
speaker-specific).
This question was studied in \cite{adi2018reverse}, in which the authors
conclude that both approaches only provide minor improvements in terms of
ASR performance.
Their speaker classification experiments also show that the baseline system
already tends to learn speaker-invariant features. However, they did not
run speaker verification experiments and hence did not assess the suitability
of these features for the goal of anonymization.


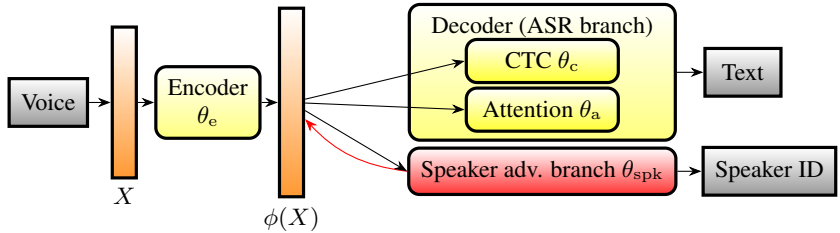
\begin{figure*}[t]
  \centering
  \begin{tikzpicture}[>=Stealth, node font=\small, scale=1,
  outernode/.style={draw, shape=rectangle, rounded corners=4pt, top color=white, bottom color=yellow!50,very thick, inner sep=.5em, minimum size=3em},
  innernode/.style={draw, shape=rectangle, rounded corners=4pt, top color=white, bottom color=yellow!80, very thick, inner sep=.5em, minimum size=1em},
  adv/.style={draw, shape=rectangle, rounded corners=4pt, top color=white, bottom color=red!80, very thick, inner sep=.5em, minimum size=1em},
  vect/.style={draw, shape=rectangle, top color=white, bottom color=orange!80, very thick, inner sep=.5em, minimum size=3em},
  reversal/.style={draw, shape=rectangle, top color=white, bottom color=red!80,  inner sep=.5em, minimum size=3em},
  inout/.style={draw, shape=rectangle, top color=white, bottom color=gray!80, very thick, inner sep=.5em, minimum size=2em}]
  \node[inout] (voice) at (-2,.5) {Voice};
  \node[vect, minimum width=.2cm, minimum height=2cm, label=below:$X$]  (input) at (-1,0.5) {};
  \node[outernode, align=center] (encoder) at (0.1,.5) {Encoder\\ $\theta_\enc$};
  \node[vect, minimum width=.2cm, minimum height=2.5cm, label=below:$\phi(X)$]  (repres) at (1.2,0.5) {};
  \node[outernode, align=center, minimum width=3.5cm] (decoder) at (4.5,.9) {Decoder (ASR branch)\\[.9cm]};
  \node[innernode, align=center,minimum width=2cm] (ctc) at (4.5,1.05) {CTC $\theta_\ctc$};
  \node[innernode, align=center,minimum width=2cm, below=.1em of ctc] (att) {Attention $\theta_\att$};
  \node[adv, align=center,  minimum width=2cm, below=.2em of decoder] (spk)   {Speaker adv. branch $\theta_\spk$};
  \node[inout, minimum width=1cm, right=1.1em of decoder] (text)  {Text};
  \node[inout, right=1em of spk] (spkid)  {Speaker ID};
  \draw (voice) edge[->] (input.west);
  \draw (input) edge[->] (encoder.west);
  \draw (encoder) edge[->] (repres.west);
  \draw (repres) edge[->] (att.west);
  \draw (repres) edge[->] (ctc.west);
  \draw (repres) edge[<-, color=red, bend left=-20] (spk.west);
  \draw (repres) edge[->] (spk.west);
  \draw (decoder.east) edge[->] (text.west);
  \draw (spk) edge[->] (spkid.west);
\end{tikzpicture}
\caption{Architecture of the proposed model. The speaker-adversarial
  branch is shown as a red box. The red arrow indicates {\it gradient
  reversal}. When the model is deployed, the encoder could reside at the
  client side, while the decoder can be hosted by cloud services.}
\label{fig:arch}
\vspace{-1.5em}
\end{figure*}

In contrast to these studies which aim to increase ASR performance, our goal is to assess the potential benefit of adversarial training for concealing speaker identity in the context of privacy-friendly ASR. Our contributions are the following. First, we combine CTC, attention and adversarial learning within an end-to-end ASR framework. Second, we design a rigorous protocol to quantify speaker identity in ASR representations through a series of closed-set classification and open-set verification experiments. Third, we run these experiments on the Librispeech corpus \cite{panayotov2015librispeech} and show that this framework dramatically reduces speaker classification accuracy, but does not increase speaker verification error. We suggest several possible reasons behind this disparity.

The structure of the rest of the paper is as follows.
In Section~\ref{sec:arch}, we describe the baseline ASR model and our proposed
adversarial model.
Section~\ref{sec:exp} explains the experimental setup and presents our
results.
Finally, we conclude and discuss future work in Section~\ref{sec:conc}.

\section{Proposed model}
\label{sec:arch}


We start by describing the ASR model we use as a baseline, before introducing
our speaker-adversarial network.

\subsection{Baseline ASR model}

We use the end-to-end ASR framework presented in \cite{watanabe2017hybrid}
as the baseline architecture. It is composed of three sub-networks: an 
\emph{encoder} which transforms the input sequence of speech feature vectors into a new
representation $\phi$, and two \emph{decoders} that
predict the character sequence from $\phi$. We assume that these networks have already been trained
using data
previously collected by the service provider (which may be public data,
opt-in user data, etc). Then, in the deployment phase of the system
that we envision, the encoder would run on the user device and the
resulting representation $\phi$ would be sent to the cloud for decoding.%

The first decoder is based on CTC and
the second on an attention mechanism. As argued in \cite{watanabe2017hybrid},
attention works well in most cases because it does not assume
conditional independence between the output labels (unlike CTC). However, it
is so flexible that it allows nonsequential alignments which are undesirable
in the case of ASR. Hence, CTC acts as a regularizer to prune
such misaligned hypotheses.
We denote by $\theta_{\enc}$ the parameters of the encoder,
and by $\theta_{\ctc}$ and $\theta_{\att}$ the parameters of the
CTC and attention decoders respectively. The model is trained in an
end-to-end fashion by
minimizing an objective function $\mathcal L_{\asr}$ which is a combination of
the losses $\mathcal{L}_c$ and $\mathcal{L}_a$ from both decoder branches:
$$\min_{\theta_{\enc},\theta_{\ctc},\theta_
{\att}} \mathcal L_{\asr}(\theta_{\enc},\theta_{\ctc},\theta_{\att}) =
\lambda
\mathcal{L}_c(\theta_{\enc},\theta_{\ctc}) + (1-\lambda)\mathcal{L}_a(\theta_
{\enc},\theta_{\att}),$$
with $\lambda\in[0,1]$ a trade-off parameter between the two
decoders.

We now formally describe the form of the two losses $\mathcal{L}_c$
and $\mathcal{L}_a$. We denote each sample in the dataset as $S_i = (X_i, Y_i, z_i)$,
where $X_i = \{x_1, ... ,x_T\}$ is the sequence of $T$ acoustic feature
frames, $Y_i = \{y_1, ... , y_M\}$ is the sequence of $M$ characters in
the transcription, and $z_i$ is the speaker label. In the case of CTC, several
intermediate label sequences of length $T$ are created by repeating characters and
inserting a speacial {\it blank} label to mark character boundaries. Let $\Psi(Y_i)$ be the set of
all such intermediate label sequences. The CTC loss $
\mathcal{L}_{c}(\theta_{\enc},\theta_{\ctc})$ is computed as $\mathcal{L}_{c} = -\ln P(Y_i | X_i; \theta_{\enc},\theta_{\ctc})$
where $P(Y_i | X_i ;
\theta_{\enc},\theta_{\ctc}) = \sum_{\psi \in \Psi(Y_i)} P(\psi | X_i ;
\theta_{\enc},\theta_{\ctc})$. This sum is computed by assuming conditional
independence over $X_i$, hence $ P(\psi | X_i ; \theta_{\enc},\theta_{\ctc}) = \prod_{t = 1}^{T} P(\psi_t
| X_i; \theta_{\enc},\theta_{\ctc}) \approx \prod_{t = 1}^{T} P(\psi_t; \theta_{\enc},\theta_{\ctc})$.
The attention branch does not require an intermediate label representation and
conditional independence is not assumed, hence the loss is simply computed as
$\mathcal{L}_{a}(\theta_{\enc},\theta_{\att}) = - \sum_{m \in M} \ln P(y_m |
X_i, y_{1:m-1} ; \theta_{\enc},\theta_{\att})$.

\subsection{Speaker-adversarial model}

In order to encourage the network to learn representations that are not only
good at ASR but also hide speaker identity, we propose to extend the above
architecture with what we call a \emph{speaker-adversarial} branch.
This branch models an adversary which attempts to infer the
speaker identity from the encoded representation $\phi$. We denote by
$\theta_s$ the parameters of the speaker-adversarial branch. Given the
encoder parameters $\theta_{\enc}$, the goal of the
adversary is to find $\theta_{s}$ that minimizes the loss $\mathcal{L}_
{\spk}(\theta_{\enc},\theta_s) = -\ln P (z_i | X_i ; \theta_{\enc},\theta_s)$.
Our new model
is then trained in an end-to-end manner by optimizing the following min-max
objective:
\begin{equation*}
\min_{\theta_{\enc},\theta_{\ctc},\theta_
{\att}}\max_{\theta_s}~
\mathcal{L}_\asr(\theta_{\enc},\theta_{\ctc},\theta_
{\att}) -\alpha \mathcal L_\spk(\theta_{\enc},\theta_s),
\end{equation*} 
where $\alpha\geq 0$ is a trade-off parameter between the ASR objective and
the
speaker-adversarial objective. The baseline network can be recovered by setting
$\alpha=0$. Note that the max part of the objective corresponds to the
adversary, which controls only the speaker-adversarial parameters
$\theta_s$.
The goal of the speaker-adversarial branch is to act as a
``good adversary'' and produce useful gradients to remove the speaker identity
information from the encoded representation $\phi$.
In practice, we use a \emph{gradient reversal
layer} \cite{ganin2016domain} between the encoder and the speaker-adversarial
branch so that the whole network can be trained end-to-end via
backpropagation. We refer to Fig.~\ref{fig:arch} for an illustration of the
full
architecture.

\section{Experimental evaluation}
\label{sec:exp}

\subsection{Datasets}
We use the Librispeech corpus~\cite{panayotov2015librispeech} for all the
experiments. We use different subsets for ASR training,
adversarial training, and speaker verification. For the sake of clarity we
refer to them as {\it data-full}, {\it data-adv}, and {\it data-spkv}, 
respectively (see Table~\ref{tab:data}). The {\it data-full} set is almost the
original Librispeech corpus, including {\it train-960} for training, {\it dev-clean} and {\it dev-other} for validation, and {\it test-clean} and {\it test-other} for test, except that utterances with more than 3,000 frames or more than 400 characters have been removed from {\it train-960} for faster training.

The {\it data-adv} set is a 100~h subset of {\it train-960}, which is obtained by removing long utterances from the original Librispeech {\it train-100} set similarly to above. It is split into three subsets
in order to perform closed-set speaker identification experiments, since the
speakers in the original train/dev/test splits are disjoint. There are 251
speakers in {\it data-adv}: we assign 2 utterances per speaker to each {\it
test-adv} and {\it dev-adv}. The remaining utterances are used for training and referred to as {\it train-adv}.

For speaker verification with x-vectors \cite{snyder2018x}, we use
{\it data-spkv}, which is again derived from {\it data-full}. The {\it
train-960} subset was augmented using room impulse responses, isotropic and
point-source noises~\cite{ko2017study} as well as music and speech~\cite{musan2015} as per the standard {\it sre16} recipe for training x-vectors
\cite{snyder2018x} from the Kaldi toolkit \cite{povey2011kaldi}, which we
adapted to Librispeech. This
increased the amount of data by a factor of 4.
A subset of the augmented data containing 373,985 utterances
was used to train the x-vector representation and another subset containing 422,491
utterances to train the probabilistic linear discriminant analysis (PLDA) backend. These
subsets are referred
to as {\it train-spkv} and {\it train-plda}, respectively. For evaluation, we
built an enrollment set ({\it test-clean-enroll}) and a trial set ({\it test-clean-trial}) from the {\it test-clean} data. Out of 40, 29 speakers were selected from {\it test-clean} based on sufficient data availability. For each speaker, we selected a 1~min subset after speech activity detection for enrollment and used the rest for trials. The details of the trials are
given in Table~\ref{tab:trial-data}.

\begin{table}[h]
\caption{Splits of Librispeech used in our experiments.}
\label{tab:data}
\centering
\begin{tabular}{cccc}
\hline
\textbf{dataset} & \textbf{data split} & \textbf{\# utts} & \textbf{duration (h)} \\ \hline
\multirow{5}{*}{\textit{data-full}} & train-960 & 281,231 & 960.98 \\
 & test-clean & 2,620 & 5.40 \\
 & dev-clean & 2,703 & 5.39 \\
 & test-other & 2,939 & 5.34 \\
 & dev-other & 2,864 & 5.12 \\ \hline
\multirow{3}{*}{\textit{data-adv}} & train-adv & 27,535 & 97.05 \\
 & dev-adv & 502 & 1.77 \\
 & test-adv & 502 & 1.77 \\ \hline
\multirow{4}{*}{\textit{data-spkv}} & train-spkv & 373,985 & 1,388.79 \\
 & train-plda & 422,491 & 1,443.96 \\
 & test-clean-enroll & 438 & 0.75 \\
 & test-clean-trial & 1496 & 3.60 \\ \hline
\end{tabular}
\vspace{-2em}
\end{table}

\begin{table}[th]
\centering
\caption{Detailed description of the trial set (test-clean-trial) for speaker
verification experiments.}
\label{tab:trial-data}
\begin{tabular}{ccc}
\hline
                         & \textbf{Male} & \textbf{Female} \\ \hline
\textbf{\# Speakers}      & 13            & 16              \\
\textbf{\# Genuine trials}  & 449           & 548             \\
\textbf{\# Impostor trials} & 9,457          & 11,196           \\ \hline
\end{tabular}
\vspace{-2em}
\end{table}

\subsection{Evaluation metrics}
For all tested systems, we measure ASR performance in terms of
the word
error rate (WER) and we assess the amount of information about speaker identity
in the encoded speech representation
in terms of both speaker classification
accuracy (ACC) and speaker verification equal error rate (EER).
The WER is reported on the {\it test-clean} set.
The ACC measures how well speakers can be discriminated in a closed-set setting, i.e., speakers are
known at training time.
It is evaluated over the {\it test-adv} set using the same classifier architecture 
as the speaker-adversarial branch of the proposed model (see
Section~\ref{sec:arch}).
As opposed to the ACC, the EER
measures how well the representations hide the speaker identity for unknown
speakers, in an open-set scenario. It reflects the process of confirming whether
a person is actually who the attacker thinks it might be. It evaluated over the trial set  (see Table~\ref{tab:trial-data}) using x-vector-PLDA.


The ACC and the EER will be computed for different representations: the baseline
filterbank features, the representations encoded by
the network trained for ASR only
(corresponding to $\phi_0$) as well as those obtained with the
speaker-adversarial approach (corresponding to $\phi_\alpha$ for some values
of $\alpha>0$).


\begin{figure*}[htb]
  \centering
  \subfigure[Filterbank]{\label{fig:tsnebaseline}\includegraphics[width=.24\linewidth,trim={2cm 1.3cm 1.5cm 1.3cm},clip]{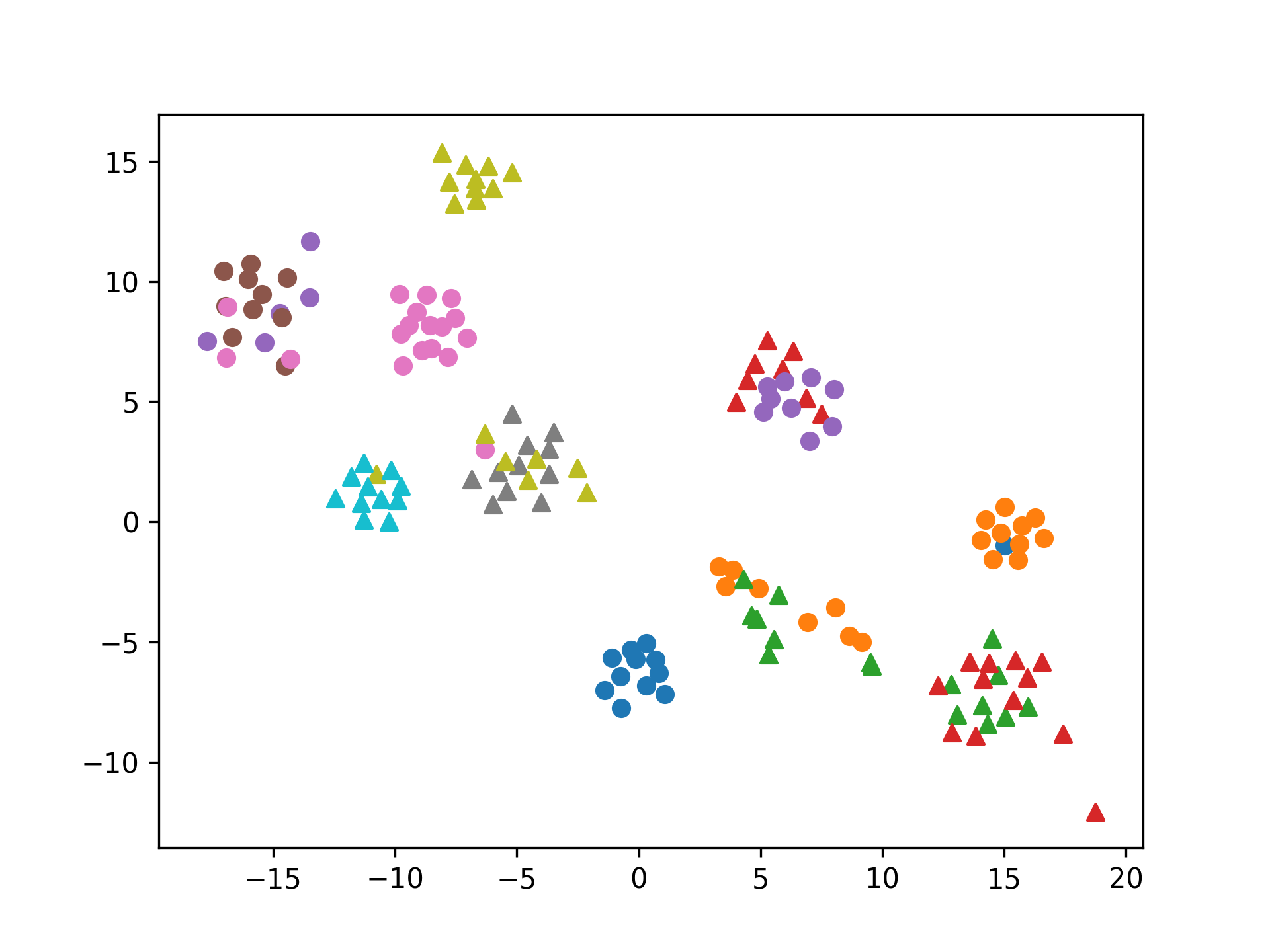}}
  \subfigure[$\alpha=0$]{\label{fig:tsne0}\includegraphics[width=.24\linewidth,trim={2cm 1.3cm 1.5cm 1.3cm},clip]{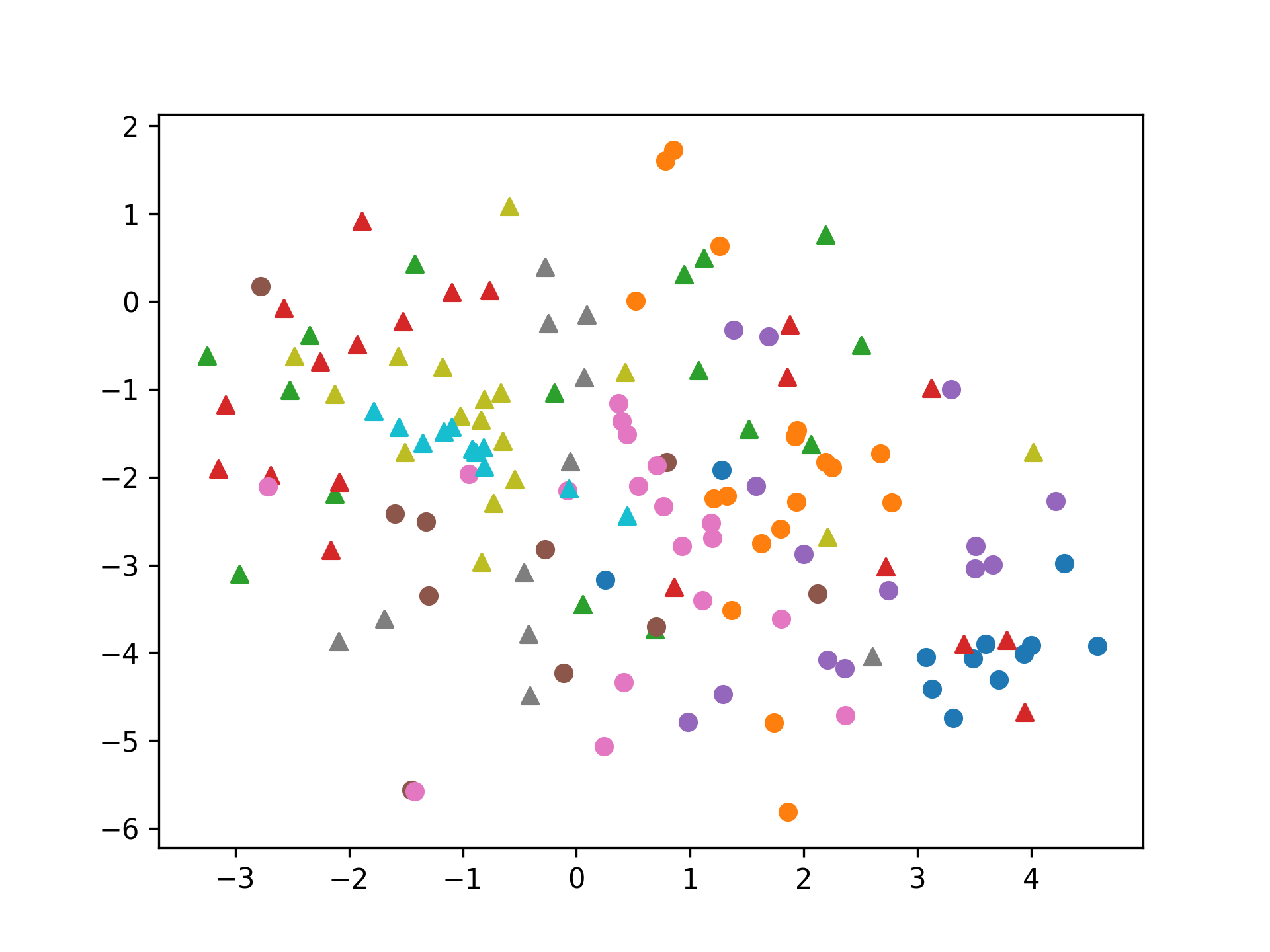}}
  \subfigure[$\alpha=0.5$]{\label{fig:tsne5}\includegraphics[width=.24\linewidth,trim={2cm 1.3cm 1.5cm 1.3cm},clip]{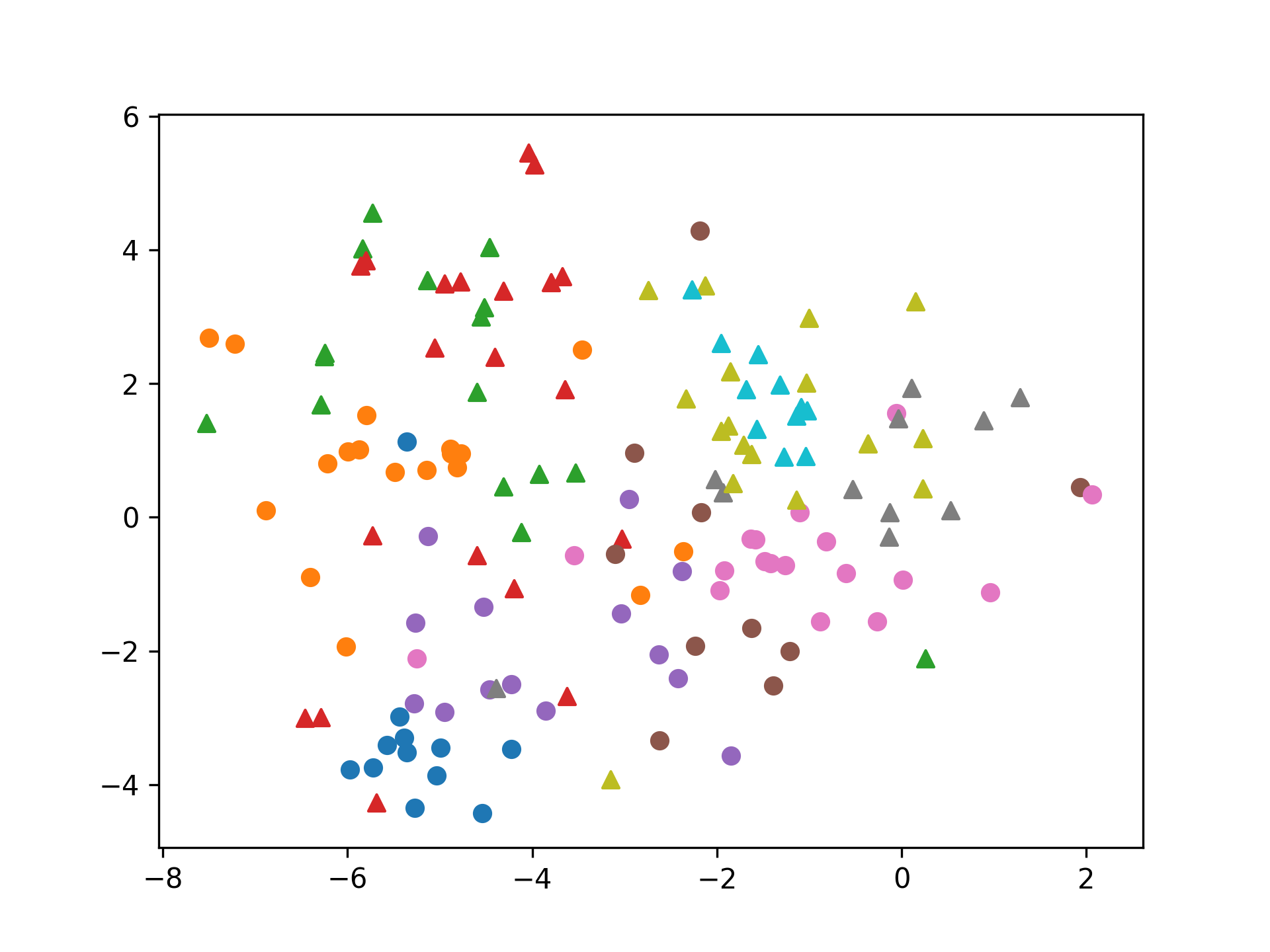}}
  \subfigure[$\alpha=2$]{\label{fig:tsne2}\includegraphics[width=.24\linewidth,trim={2cm 1.3cm 1.5cm 1.3cm},clip]{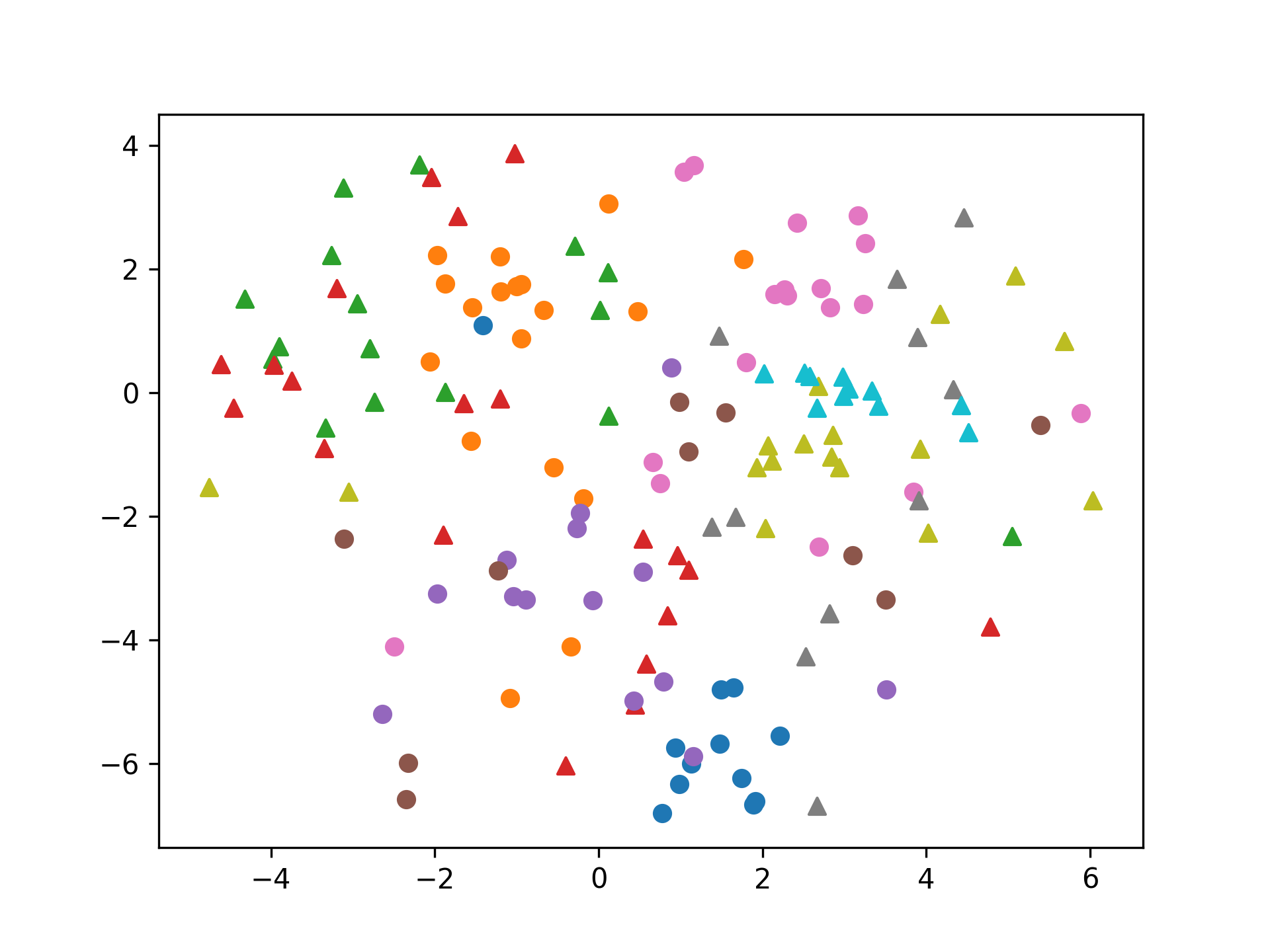}}
  \caption{Visualization of x-vector representations of 20 utterances
  of 10 speakers computed by t-SNE (perplexity equals to
  30). Males are represented by circles and females by triangles.}
  \label{fig:tsne}
  \vspace{-1.5em}
\end{figure*}

\subsection{Network architecture and training}
For all experiments, we use the ESPnet~\cite{watanabe2018espnet} toolkit which
implements the hybrid CTC/attention architecture~\cite{watanabe2017hybrid}.
The input features are 80-dimensional mel-scale filterbank coefficients with pitch and energy features, totalling 84 features per frame. 
The {\it encoder} is composed of a VGG-like convolutional neural network (CNN) layer followed by 5 bidirectional long short-term memory (LSTM) layers with 1,024 units. The VGG layer contains 4 convolutional layers followed by max pooling. The feature maps used in the convolution layers are of dimensions $(1 \times 64)$, $(64 \times 64)$, $(64 \times 128)$ and $(128 \times 128)$.
The attention-based decoder consists of location-aware attention~\cite{chorowski2015attention} with 10 convolutional channels of size
100 each followed by 2 LSTM layers with 1,024 units.
The CTC loss is computed over several possible
label sequences using dynamic programming. In all experiments, the trade-off
parameter $\lambda$ between the two decoder losses is set to 0.5. We train a single-layer recurrent
neural network language model (RNNLM) with 1,024 hidden units
over the {\it train-960} transcriptions and use it to rescore the
ASR hypotheses. The resulting WER is very close to the state of the art \cite{zeghidour2018fully} when trained on {\it train-960}.
Finally, we implemented the {\it speaker-adversarial} branch via 
a 3 bidirectional LSTM layers with 512 units followed by a softmax layer with 251 outputs corresponding to the 251 speakers in \emph{data-adv}. The adversarial loss $\mathcal{L}_{\spk}$ is summed across all vectors in the sequence. The speaker label $z_i$ is duplicated to match the length of the sequence, which is smaller than $T$ due to the subsampling performed within the encoder.
Due to this subsampling as well as to the use of bidirectional LSTM layers within the encoder and the {\it speaker-adversarial} branch, the frame-level adversarial loss approximates well a utterance-level speaker loss that would be computed from a fixed-sized utterance-level representation, while being easier to train.

In all experiments, we start by pre-training the ASR branch for
10 epochs over
\emph{data-full} and then the speaker-adversarial
branch for 15 epochs on \emph{data-adv} in order to get a strong
adversary on the pre-trained encoded representations.
Then, due to time constraints, all networks are fine-tuned on 
\emph{data-adv}: we run 15 epochs of adversarial training (which
corresponds to simple ASR
training when $\alpha=0$). Due to this, the WER is comparable to that typically achieved by end-to-end methods when trained on the {\it train-100} subset of Librispeech rather than 
the full {\it train-960} set.
Finally, freezing the resulting encoder, we further fine-tune the
speaker-adversarial branch only for 5 epochs to make sure that the reported
ACC reflects the performance of a well-trained adversary.

The {\it encoder} network contains 133.5M parameters. To encode a 10s audio file, it perform 1.1e12 arithmetic operations which can be executed in-parallel on a 40 core CPU in 17.6s and on a single Tesla P100 GPU in 149ms.

\subsection{Results and discussion}
\label{sec:spk-embedding}
We train our speaker-adversarial network for $\alpha \in \{0, 0.5, 2.0\}$,
leading to three encoded representations $\phi_\alpha
(X)$. Recall that $\alpha=0$ corresponds to the baseline ASR system as it
ignores the speaker-adversarial branch. Table~\ref{tab:results} summarizes
the results.

The first column presents the ACC and EER obtained with the input filerbank features, which are consistent with the numbers reported in the literature.
As expected, speaker
identification and verification can be addressed to very high accuracy on those features.
Using the encoded representation $\phi_0
(X)$ trained for ASR only already provides a significant privacy gain: the ACC
is divided by 2 and
the EER is multiplied by 4, which suggests that a reasonable amount of speaker information is removed during ASR training. Nevertheless, $\phi_0
(X)$ still contains some speaker identity information. 

More
interestingly, our results clearly show that adversarial training
drastically reduces the performance in speaker identification but not in
verification. On the other hand, and counterintuitive to the speaker-invariance
claims by several previous studies, we observe that the verification
performance actually improves after adversarial training. This exhibits a
possible limitation in the generalization of adversarial training to unseen
speakers and hence establishes the need for further investigation. The reason
for the disparity between classification and verification performance might be
that the speaker-adversarial branch does not inherently perform verification and hence is not optimized for that task. It might also be attributed to the representation capacity of that branch, to the number of speakers presented during adversarial training, and/or to the exact range of $\alpha$ needed for generalizable anonymization. These factors of variation open several venues for future experiments.

\begin{table}[th]
\centering
\caption{ASR and speaker recognition results with different representations.
WER (\%) is reported on {\it test-clean} set, ACC (\%) on {\it
test-adv} set and EER (\%) on {\it test-clean-trial}.}
\label{tab:results}
\begin{tabular}{ccccc}
\hline
 & Filterbank & \textbf{$\phi_0$} & \textbf{$\phi_{0.5}$} & \textbf{$\phi_{2.0}$} \\ \hline
\textbf{WER} & -- & 10.9 & 12.5 & 12.5 \\ \hline
\textbf{ACC} & 93.1 & 46.3 & 6.4 & 2.5 \\ \hline
\textbf{EER} Pooled & 5.72 & 23.07 & 21.97 & 19.56 \\ 
\textbf{EER} Male & 3.34 & 19.38 & 18.26 & 16.26 \\ 
\textbf{EER} Female & 7.48 & 26.46 & 24.45 & 22.45 \\ \hline
\end{tabular}
\end{table}

We also notice that the WER stays reasonably low and stabilizes to the value of 12.5\% after increasing $\alpha$ from 0.5 to 2.
In particular, for $\alpha=2$ the WER is just 1.6\% absolute more than the baseline ($\alpha=0$).

We evaluate whether utterances from the same speaker stay in the same
neighborhood or are scattered in the representation space. We compute t-SNE
embeddings on the x-vector representations of 20 utterances for 10 speakers (5
male, 5 female), shown in Figure~\ref{fig:tsne}. When using filterbanks, we
can observe well-clustered utterances. The clusters break down when training
the x-vectors on $\phi_0$. For the x-vectors trained on $\phi_{0.5}$ and
$\phi_{2.0}$, the clusters start to re-emerge. The silhouette scores for x-vectors extracted from filterbank, $\phi_0$, $\phi_{0.5}$ and $\phi_{2.0}$ representations are $0.14$, $-0.17$, $-0.05$ and $-0.09$ respectively, are consistent with the observed EER values.


\section{Conclusions and future work}
\label{sec:conc}

We proposed to combine CTC and attention losses with a speaker-adversarial loss within an end-to-end framework with the goal of learning privacy-preserving representations for ASR. Such representations could be safely transmitted to cloud-services for decoding. We investigate the level of speaker identity anonymization achieved by adversarial training through closed-set speaker classification and open-set speaker verification measures. Adversarial training appears to dramatically reduce the closed-set classification accuracy, seemingly indicating a high-level of anonymization. However, this observation does not match with the open-set verification results, which correspond to the real scenario of an adversary trying to confirm the identity of a suspected speaker. Hence we conclude that the adversarial training does not immediately generalize to produce anonymous representations in speech. We hypothesize that this disparity might be attributed to the representation capacity of the adversarial branch, the size of the training set, the formulation of the adversarial loss, and/or the value of the trade-off parameter with the ASR loss. As a future work, we plan to modify the speaker adversarial branch to inherently optimize for verification instead of classification and ascertain the impact of these experimental choices over different datasets, including for languages not seen in training.

\section{Acknowledgements}
\label{sec:ack}

This work was supported in part by the European Union's Horizon 2020 
Research and Innovation Program under Grant Agreement No. 825081 
COMPRISE (\url{https://project.inria.fr/comprise/}) and by the French 
National Research Agency under project DEEP-PRIVACY 
(ANR-18-CE23-0018). Experiments were carried out 
using the Grid'5000 testbed, supported by a scientific interest group 
hosted by Inria and including CNRS, RENATER and several Universities as 
well as other organizations. The 
authors would like to thank Md Sahidullah for providing the speaker verification data split.

\bibliographystyle{IEEEtran}
\bibliography{mybib}

\end{document}